\documentclass{article}

\usepackage[final,nonatbib]{neurips_2019}
\usepackage{bm}
\usepackage{graphicx} 
\usepackage[utf8]{inputenc} 
\usepackage[T1]{fontenc}    
\usepackage{hyperref}       
\usepackage{url}            
\usepackage{booktabs}       
\usepackage{amsfonts}       
\usepackage{nicefrac}       
\usepackage{microtype}      
\usepackage{amsmath}
\usepackage{authblk}
\usepackage{algorithm}
\usepackage{amsmath,amssymb,amsthm}
\usepackage{algpseudocode}
\usepackage{multirow}
\usepackage{lineno}
\usepackage{subfig}
\usepackage{wrapfig}

\def\bx{\mathbf x}

\def\by{\mathbf y}

\def\bW{\mathbf W}

\def\bOne{\mathbf 1}
\def\btheta{\boldsymbol \theta}
\def\bvartheta{\boldsymbol \vartheta}
\def\bydef{\triangleq}
\usepackage{accents}
\newcommand{\ubar}[1]{\underaccent{\bar}{#1}}

\newcommand{\T}{\scriptscriptstyle T}

\title{Learn Electronic Health Records by Fully Decentralized Federated Learning}

\author[1]{Songtao Lu}
\author[2]{Yawen Zhang}
\author[2]{Yunlong Wang\thanks{Email: yunlong.wang@iqvia.com}}
\author[3]{Christina Mack}

\affil[1]{Department of Electric and Computer Engineering, University of Minnesota}
\affil[2]{Advanced Analytic, IQVIA}
\affil[3]{Epidemiology and Clinical Evidence, IQVIA}

\begin{document}
\maketitle
\begin{abstract}
Federated learning opens a number of research opportunities due to its high communication efficiency in distributed training problems within  a star network. In this paper, we focus on improving the communication efficiency for fully decentralized federated learning over a graph, where the algorithm performs local updates for several iterations and then enables communications among the nodes. In such a way, the communication rounds of exchanging the common interest of parameters can be saved  significantly without loss of optimality of the solutions. Multiple numerical simulations based on large, real-world electronic health record databases showcase the superiority of the decentralized federated learning compared with classic methods.
\end{abstract}

\section{Introduction}
In this era of big data, the use of aggregated patient information can effectively train a high-quality machine learning model by adopting multiple computational resources. However, there are several challenges to this exercise. First, data privacy and security are paramount, and they are often difficult to integrate the data collected and aggregated across. Second, communication efficiency presents a challenge, as each communication round may result in long delays especially in applications of the internet of things (IoT) or self-driving systems. To overcome these challenges, federated learning may be an    effective way to increase training efficiency and allow knowledge to be shared without compromising user privacy\cite{brisimi2018federated}.


\subsection{Motivation}

Decentralized federated-learning techniques are promising across numerous applications, such as smart healthcare, etc. Medical data such as disease symptoms and medical recordings are highly sensitive, and collecting clinical datasets from isolated medical centers and hospitals is a challenge. Federated learning, enabling multiple agents collaboratively learn a shared prediction model while keeping all the training data private, could play a pivotal role in solving this problem \cite{mcma17,yang2019federated}. 

Patient data is fully decentralized in most real-world applications and  patient-level data exchange among stakeholders such as insurance companies and treating facilities  is prohibited by laws, such as the United States Health Insurance Portability and Accountability Act (HIPPA) \cite{act1996health}. Therefore,  hospitals have hundreds of patient-level records in one disease area, describing the characteristics of every patient, but lack the breadth of the information about the patients; with these limited samples, a complex model cannot be trained by one hospital.

Nevertheless, under an agreement, each hospital is allowed to share non-sensitive intermediate statistics that are strictly de-identified  and aggregated \cite{toh2013confounding, toh2014privacy, brown2010distributed}. In this setting, the hospitals constitute an undirected network where each hospital is a node, and an only neighboring node can exchange information. Also, note that the data are non-identical, since the hospitals are located in different areas and the environmental factors have much impact on people's health status. We will show in this paper that by implementing decentralized iterative optimization algorithm, every node will reach the consensus optimality as if it owns all the data as a fictitious fusion center. Here we would like to emphasize that the studied application is decentralized rather than distributed with a star network, as it is infeasible to have a fusion center that is trusted by every node to collect healthcare data.

 \vspace{-8px}
\subsection{Scope of This Work}
 \vspace{-8px}
In practice, transmitting messages over networks requires much more effort and spending resources compared with local computation, such as encryption, coding/decoding, channel equalization, etc. Therefore, it is of interest of performing local update to learn the models. The current federated learning strategies are mainly performed over a star network  \cite{li2014co,mcma17,yu2019linear} through applying the traditional distributed optimization algorithms, such as distributed (stochastic) gradient descent \cite{nedic09,ram10}. By adopting a central controller or parameter server, the slave nodes implement multiple rounds of local updates and then communicate with the master node such that a large amount of the communication rounds among the nodes can be saved.  It has been shown in \cite{yu2019linear} that there are only $\mathcal{O}((NT)^{3/4})$ number of communication rounds required instead of $\mathcal{O}(T)$ in the classic decentralized non-convex setting for the non-identical datesets, where $N$ denotes the total number of nodes and $T$ stands for the total number of iterations.

In this work, we propose a fully decentralized federated learning framework by leveraging two classic non-convex decentralized optimization, which are decentralized stochastic gradient descent (DSGD) \cite{lian17,jia17} and decentralized stochastic gradient tracking (DSGT) (a.k.a GNSD) \cite{luzh19}. We remark that DSGT has the advantages of dealing with non-identical datasets compared with DSGD. First,  we will introduce the proposed communication efficient decentralized training algorithm for federated learning. Then, we show the linear speedup of DSGT by quantifying its convergence rate to the first-order stationary points theoretically. Third, we numerically compare the decentralized federated learning algorithms with the classic counterparts which do not consider communication efficiency. To the best of our knowledge, this is the first work that applies fully decentralized non-convex stochastic algorithms for federated learning and obtains reasonably good results for health record datasets.

\vspace{-10px}
\section{Decentralized Stochastic Non-convex Federated Learning}
\label{gen_inst}
\subsection{Dataset}
\begin{figure}[t]
    \centering
    \includegraphics[width=.35\linewidth]{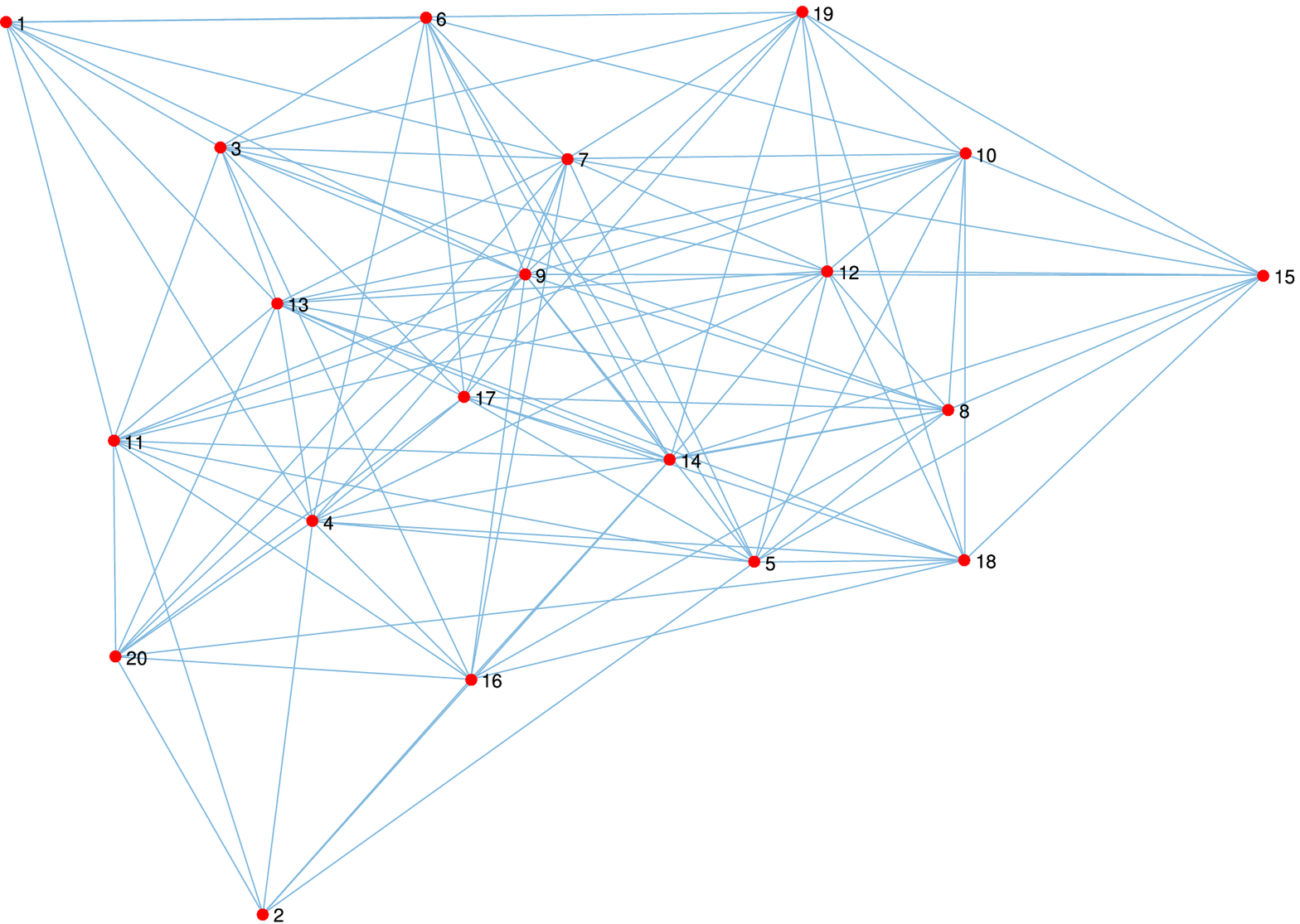}~
    \hspace{0.8cm}
    \includegraphics[width=.35\linewidth]{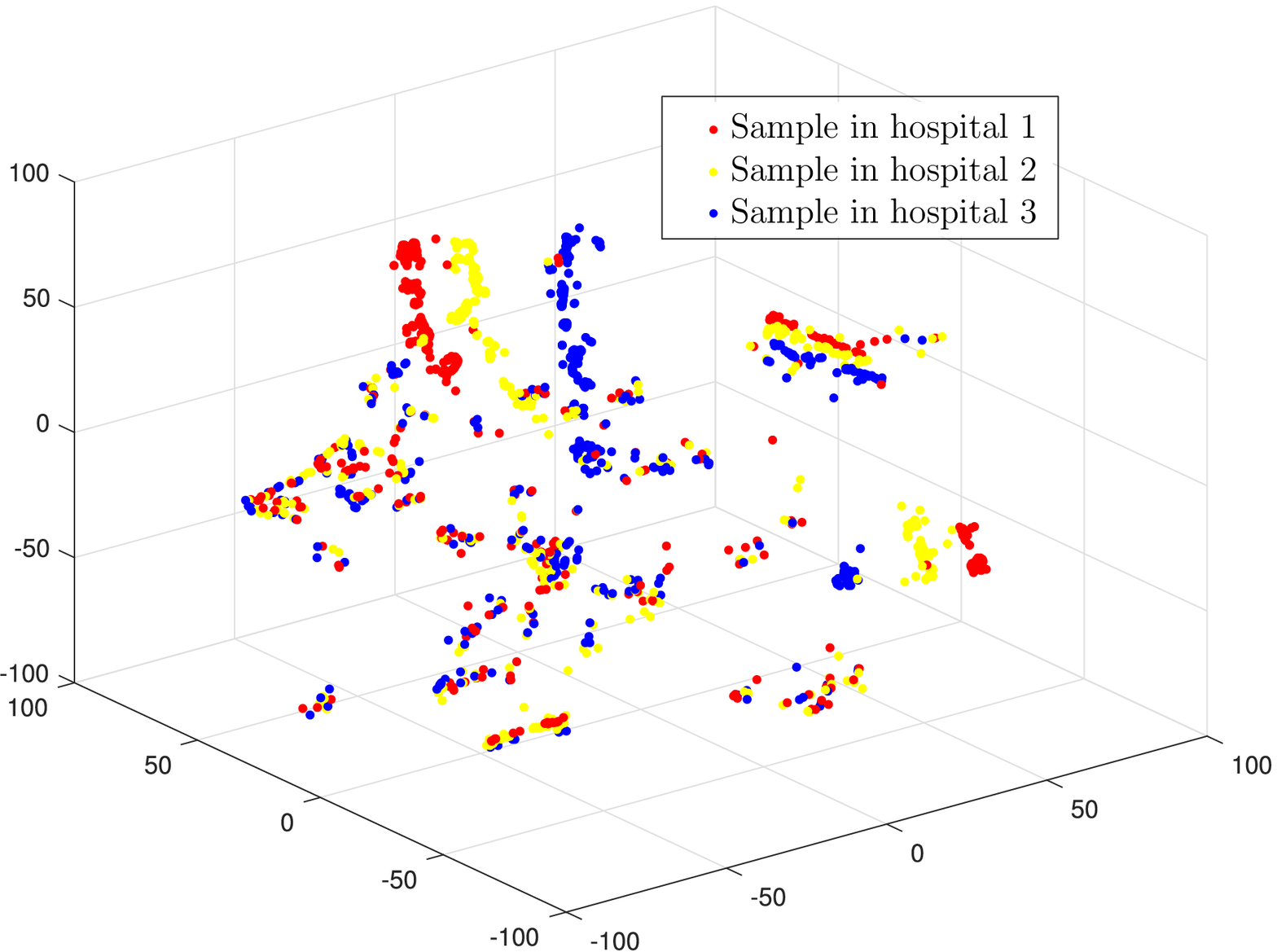}~
    \caption{\footnotesize Desription of the real health records: (Left) graph of the nodes (hospitals); (Right) t-SNE distribution of the samples in three nodes (hospitals) in the Alzheimer patients' dataset. }
    \label{fig:dataset}
    \vspace{-5mm}
\end{figure}

Data and pre-processing:
we test our algorithm on a proprietary clinical dataset that consists 2,103 patients diagnosed as Alzheimer's Disease (AD), and 7, 919 patients diagnosed with mild cognitive impairment (MCI), who have gotten early stage symptoms of AD. The electronic health records of all the patients are collected from 20 hospitals, about 500 recordings per each. The graph of all 20 hospitals is shown on the left in Fig.~\ref{fig:dataset}. And the figure on the right in Fig.~\ref{fig:dataset} gives an example of the t-SNE distribution of the samples of three hospitals. The separated distributions of different hospitals indicates the heterogeneity of the data in nature, which has been rarely addressed by previous federated system \cite{mcma17,li2019federated}. These figures motive us to develop efficient algorithms of being able to handle non-identical dataset in a decentralized setting.  

\subsection{Problem formulation}
Consider a multi-agent system that consists of $N$ agents well-connected by a graph
$\mathcal{G}\bydef\{\mathcal{V},\mathcal{E}\}$, where each of them is indexed by $i\in[N]$. The agents are capable of performing local computations and exchanging binary decisions with other agents. Each agent has a label, which is private and marked by doctors. In this work, we consider the following collaborative filtering problem, i.e.,
\vspace{-3mm}
\begin{equation}\label{eq.pro}
    \min_{\btheta_i,\forall i}\frac{1}{N}\sum^N_{i=1}f_i(\btheta_i),\quad\textrm{s.t.}\;\btheta_i=\btheta_j,\; j\in\mathcal{N}_i,\forall i
\end{equation}

where $f_i(\btheta_i)=\mathbb{E}_{\xi_i\sim\mathcal{D}_i}[F(\btheta_i,\xi_i)]$ is smooth and possibly nonconvex,  $F(\btheta_i,\xi_i)$ denotes the loss function with respect to sample $\xi_i$, $\mathcal{N}_i$ represents the set of node $i$'s neighbors, and $\mathcal{D}_i$ stands for the distribution of data at the $i$th node. Here, we consider the graph is well-connected in the sense that the following property is assumed.

{\bf Assumption 1.} \emph{Assume the weighting matrix $\bW\in\mathbb{R}^{n\times n}$ is symmetric, satisfying  
$|\ubar{\lambda}_{\max}(\bW)|<1,\quad  \bW \bOne=\bOne,
$
where $\ubar{\lambda}_{\max}(\bW)$ denotes the second largest eigenvalue of $\bW$ and $\bOne\in\mathbb{R}^{n\times 1}$ is an all one vector.} 
Problem \eqref{eq.pro} is the classic distributed optimization problem. Existing works \cite{lian17,luzh19} have shown DSGD and DSGT are able to find  an $\epsilon$-approximate first-order stationary point  in a sublinear convergence rate in the sense that the size of the gradient of the objective function and consensus violation of the iterates among all the nodes will be both small enough as the algorithm proceeds to a large number of iterations.

 \begin{wrapfigure}{R}{0.51\textwidth}
 \vspace{-5mm}
    \begin{minipage}{0.5\textwidth}
      \begin{algorithm}[H]
        \caption{{\small Fully Decentralized Non-convex Stochastic Gradient Descent for Federated Learning}}
        \label{alg:p1}
        \begin{algorithmic}
        \State {\bfseries Input:} $\btheta^{0}$, $\alpha^{0}$
        \For{$r=1,\ldots$}
        \State Randomly collect $m$ samples $\xi^r_i$ locally
        \State Calculate the stochastic gradient $\nabla g_i(\bx^r_i)$ 
        \State Each node updates $\btheta^{r+1}_i$ individually by  \eqref{eq.local}
        \If{$r$ is a multiple of $Q$, i.e.,$\mod(r,Q)=0$}
        \State Update $\btheta^{r+1}_i$ by \eqref{eq.updateofalg} or by \eqref{eq.dsgupdate}
        \EndIf
        \EndFor
        \end{algorithmic}
        \end{algorithm}
    \end{minipage}
 \vspace{-2mm}
  \end{wrapfigure}
 \vspace{-8px}
\subsection{Fully Decentralized Non-convex Stochastic Algorithm for Federated Learning}
 \vspace{-8px}
The decentralized optimization algorithms have two key steps: 1) local update 2) communications among nodes. In the federated setting we perform local update multiple times instead of one.
\vspace{-2px}
\subsubsection{Algorithm Description}
First, let $\nabla g_i(\btheta_i)=m^{-1}\sum^m_{l=1}\nabla f_i(\btheta_i,\xi_l)$, which serves as an estimate of the true gradient at each node. {\bf DSGD}: the traditional DSGD basically performs the gradient update and  communications at each step, i.e.,
 \begin{equation}\label{eq.dsgupdate}
     \btheta^{r+1}_i=\sum_{j\in\mathcal{N}_i}\bW_{ij}\btheta^r_j-\alpha^r\nabla_{\btheta_i} g_i(\btheta^{r}_i).
 \end{equation}
 
{\bf DSGT} in practice, the data is heterogeneously/non-identically distributed and the loss function is highly non-convex such as in neural networks, the most efficient/advanced decentralized algorithm is DSGT. Instead of only performing local gradient update, the update of the iterates by DSGT can be written as the following
\begin{equation}\label{eq.updateofalg}
\btheta^{r+1}_i=\sum_{j\in\mathcal{N}_i}\bW_{ij}\btheta^r_j-\alpha^r	\bvartheta^r_i,\quad \bvartheta^{r+1} =\sum_{j\in\mathcal{N}_i}\bW_{ij}\bvartheta^{r}_j+\left(\nabla_{\btheta_i} g_i(\btheta^{r+1}_i)-\nabla_{\btheta_i} g_i(\btheta^{r}_i)\right).
\end{equation}
Compared with the DSGD method, the GT technique adds some correction terms, which actually keep tracking the full gradient of the objective function so that the error terms resulted by the difference of data distributions among nodes can be shrunk quickly \cite{di2016next,sun19con}. 

Next, we introduce the decentralized federated learning as follows.

{\bf Local update}: the local update is very efficient, which only needs to compute the estimated gradient in the following way in parallel, i.e.,
 \begin{equation}\label{eq.local}
     \btheta^{r+1}_i=\btheta^r_i-\alpha^r\nabla_{\btheta_i} g_i(\btheta^{r}_i).
 \end{equation}
 Inspired by this fact, we insert the local update into the original DSGD and DSGT algorithms. The details of the algorithm is shown in Algorithm \ref{alg:p1}. It can be observed that we perform DSGD or DSGT for every $Q$ times local updates.
 \vspace{-8px}
\subsubsection{Assumptions and Properties of Algorithm}
Before showing the theoretical results, we first have the following assumptions on the problem setups.

{\bf Assumption 2.} We assume that the objective function has Lipschitz gradient continuity with constant $L$, i.e., $
\|\nabla f_i(\bx)-\nabla f_i(\by)\|\le L\|\bx-\by\|,\forall i$,  and also assume the unbiased gradient estimation
$\mathbb{E}_{\xi_i\sim\mathcal{D}_i}[\nabla_{\btheta_i} g_i(\btheta_i)]=\nabla f_i(\btheta_i),\forall i$, and bounded estimation variance
$\mathbb{E}_{\xi_i\sim\mathcal{D}_i}\|\nabla_{\btheta_i} g_i(\btheta_i)-\nabla f_i(\btheta_i)\|^2\le\sigma^2,\forall i$.

Towards this end, we also remark that relation $\bW \bOne=\bOne$  implies $\|\bW-\frac{1}{N}\bOne\bOne^{\T}\|<1$, which gives the contraction of the iterates as the algorithm iterates so that the algorithm is able to achieve the consensus quickly. With these assumptions and properties in mind, we can have the following theoretical result.

{\bf Theorem 1.} \emph{Suppose Assumption 1 and 2 hold. If we choose $\alpha^r\sim\mathcal{O}(\sqrt{N/r})$ and $Q=1$ in Algorithm 1 by adopting DSGT, then when $T$ is large we have
\begin{equation}
   \frac{1}{T}\left(\sum^T_{r=1} \left\|\frac{1}{N}\sum^N_{i=1}\nabla f_i(\btheta^r_i)\right\|^2+\frac{1}{N}\sum^N_{i=1}\|\btheta^r_i-\bar{\btheta}^r\|^2\right)\le \mathcal{O}\left(\frac{\sigma^2}{N\sqrt{T}}\right)
\end{equation}
where $\bar{\btheta}^r=1/N\sum^N_{i=1}\btheta^r_i$ denotes the average of the iterates.}

It can be observed that the optimality gap decreases in a rate of $\mathcal{O}(\sigma^2/(N\sqrt{T}))$ with a linear speedup in terms of the number of the nodes, demonstrating the key superiority of performing distributed learning over centralized one \cite{lian17,18_7519,yu2019linear}. Note that this is the first theoretical result to show the linear speedup of stochastic gradient tracking methods. Unfortunately, there is no theoretical guarantee for the case of a general $Q>1$. To the best of our knowledge, there is no any theoretical results to show the convergence of any decentralized algorithm in this setting. From the numerical results, it can be seen in the next section that the decentralized federated learning algorithm can also converge to the stationary points with much less communication rounds. 
\begin{wrapfigure}{R}{0.5\textwidth}
 \vspace{-6mm}
    \begin{minipage}{0.5\textwidth}
  \begin{center}
    \includegraphics[width=1\textwidth]{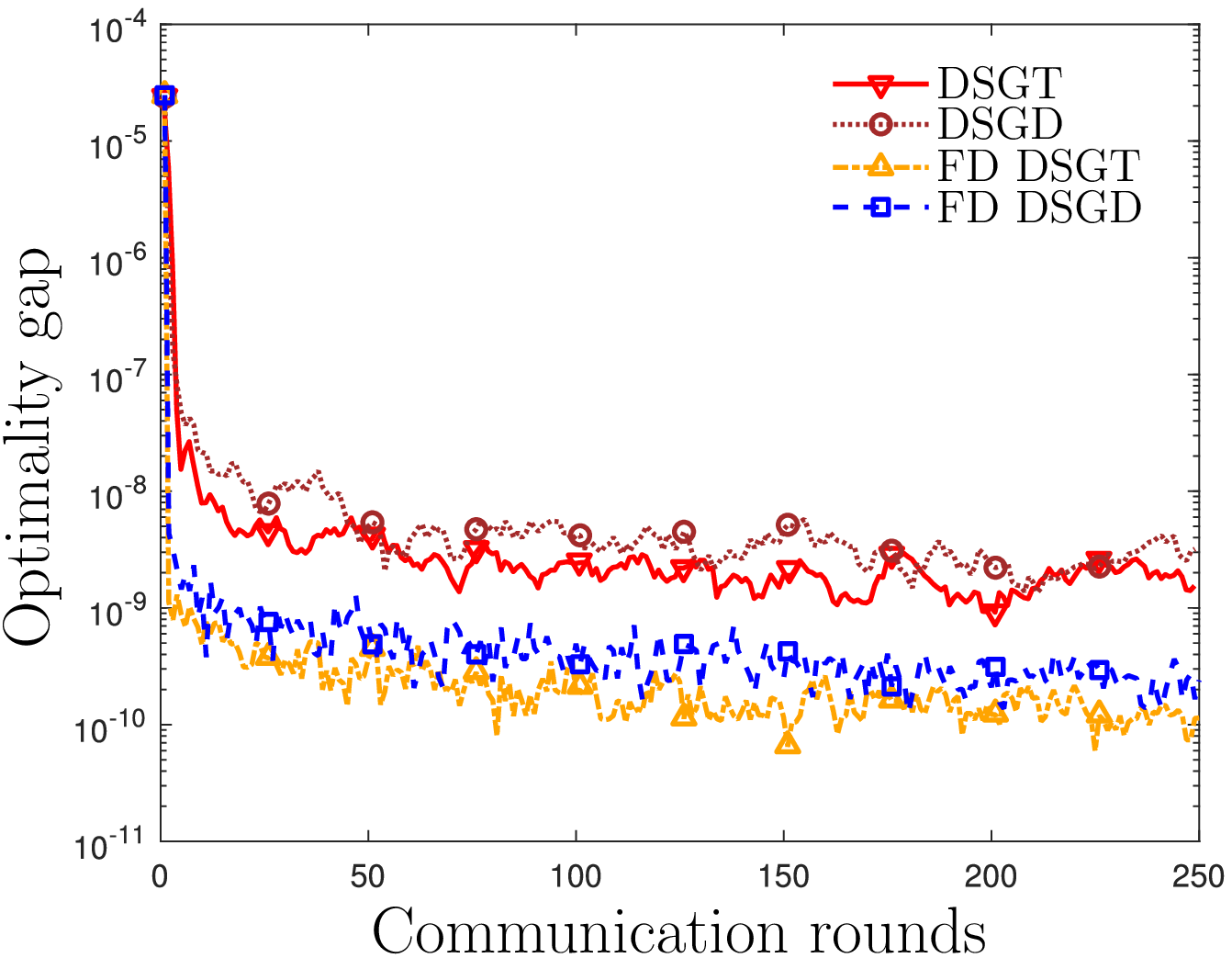}
  \end{center}
  \caption{Convergence behaviours of algorithms with respect to communication rounds}
    \end{minipage}
\vspace{-7mm}
\label{converge_fig}
\end{wrapfigure}
 \vspace{-8px}
\section{Numerical Experiments}
 \vspace{-8px}
In this section, we provide numerical results to showcase the decentralized federated learning for extracting latent features from the electronic health records. We compare DSGD, DSGT,  federated (FD) DSGD, and FD DSGT, where $m=20$, $Q=100$, $\alpha^r=0.02/\sqrt{r}$ and we train a shallow neural network at each node with a problem dimension of 42. It can be observed from Fig.~2 that FD algorithms converge much faster than classic methods in terms of communication rounds. Compared with DSGD and DSGT, DSGT in general can achieve a smaller optimality gap due to the fact the GT is able to track the full gradient while DSGD only uses the local information to update the iterates. From a theory perspective, the difference between DSGD and DSGT will be diminishing asymptotically.

\section{Concluding Remarks and Future Work}
In this work, we presented a new approach of leveraging decentralized non-convex optimization  for federated learning to extract patients features from real-world, de-identified  hospital datasets. The advantages of performing decentralized federated learning are three-fold, 1) data privacy could be preserved better than the centralized case; 2) the computational burden is released compared with the centralized processing (linear speedup), and in parallel, 3) the communication efficiency is increased. In future work, we will examine the theoretical guarantees of the algorithm for the case of $Q>1$.



\clearpage
\newpage
\bibliographystyle{IEEEtran}
\bibliography{Bibliography}

\clearpage
\newpage
\end{document}